\documentclass[conference]{IEEEtran}
\IEEEoverridecommandlockouts
\usepackage{cite}
\usepackage{amsmath,amssymb,amsfonts}
\usepackage{algorithm}
\usepackage{algorithmic}
\usepackage{graphicx}
\usepackage{textcomp}
\usepackage{xcolor}

\def\BibTeX{{\rm B\kern-.05em{\sc i\kern-.025em b}\kern-.08em
    T\kern-.1667em\lower.7ex\hbox{E}\kern-.125emX}}

\begin{document}

\title{Vulnerability of Face Recognition Systems Against Composite Face Reconstruction Attack }

\author{\IEEEauthorblockN{Hadi Mansourifar}
\IEEEauthorblockA{\textit{Computer Science Department} \\
\textit{University Of Houston}\\
Houston, USA \\
hmansourifar@uh.edu}

\and
\IEEEauthorblockN{Weidong Shi}
\IEEEauthorblockA{\textit{Computer Science Department} \\
\textit{University Of Houston}\\
Houston, USA \\
wshi3@uh.edu}
}

\newcommand{\keshav}[1]{\textcolor{red}{Keshav: {#1}}}

\maketitle
\begin{abstract}
  Rounding confidence score is considered trivial but a simple and effective countermeasure to stop gradient descent based image reconstruction attacks. However, its capability in the face of more sophisticated reconstruction attacks is an uninvestigated research area.
In this paper, we prove that, the face reconstruction attacks based on composite faces can reveal the inefficiency of rounding policy as countermeasure. 
We assume that, the attacker takes advantage of face composite parts which helps the attacker to get access to the most important features of the face or decompose it to the independent segments. Afterwards, decomposed segments are exploited as search parameters to create a search path to reconstruct optimal face. Face composition parts enable the attacker to violate the privacy of face recognition models even with a blind search. However, we assume that, the attacker may take advantage of random search to reconstruct the target face faster.
The algorithm is started with random composition of face parts as initial face and confidence score is considered as fitness value. Our experiments show that, since the rounding policy as countermeasure can't stop the random search process, current face recognition systems are extremely vulnerable against such sophisticated attacks.  To address this problem, we successfully test Face Detection Score Filtering (FDSF) as a countermeasure to protect the privacy of training data against proposed attack.

\end{abstract}

\section{Introduction}
With the rise of machine learning-as-a-service (MLaaS) [1,2]
online pre-trained models are increasingly exploited
in applications like facial recognition [3,4]. The accuracy of trained facial recognition models has been one and only priority in the past decade. In recent years however, preserving the privacy of people whose personal photo has been used as training data became a major priority. In the context of these services, an imminent threat is that, the attacker can reconstruct a recognizable image of a person, given the name of corresponding person and the confidence score returned by API.
 In this case, breaching the sensitive, confidential or protected
data is no longer considered as a financial risk, but it may lead to life threatening danger for victims. In a reconstruction attack recently introduced by Fredrikson et al. [5], adversarial access to a trained facial recognition model is abused
to reconstruct the face of an individual given the corresponding name or identifier via gradient descent optimization. However, obtained results by Fredrickson et al. [5] showed that, a reasonable rounding level can completely obviate the threat of gradient descent based attacks. Unfortunately, this trivial countermeasure could not guarantee the privacy of personal images when it comes to deal with more sophisticated attacks. 
In this paper, we show that, the attacker can impose a significant threat against face recognition systems by taking advantage of optimization techniques like random search which are not dependent to gradient descent.
 We disclose the imminent threat of face composition parts as it helps the attacker to form a state space of faces to search a composite face with maximum confidence score. As a result, rounding the confidence score as countermeasure can not stop the attacker and training data would no longer remain private.
The strength of proposed attack is that, rather than trying to reach an identical reconstructed face, it tries to synthesize an instance which resembles the target face. This policy helps the attacker to reach an approximation of a difficult problem by a nearby problem that is easier to solve.
Our observations show that, this type of attack needs more sophisticated countermeasures.
To tackle this formidable threat, we propose Face Detection Score Filtering (FDSF) as an effective countermeasure. The main idea of FDSF is to return high confidence score to face which get low Face Detection Score (FDS). The intuition behind FDSF is that, the composite faces get lower FDS comparing to real faces. Returning high confidence score to low quality faces can fool the attacker to form a wrong search path.
The experimental results show that, the proposed attack cannot be nullified by rounding policy as countermeasure. Besides, the obtained reconstructed faces significantly resemble the target faces in terms of 17 face characteristics measured by an independent face recognition system.
Our contributions are as follows.

\begin{itemize}
       \item We show the potential of composite faces to impose a significant threat against privacy of face recognition systems.
        \item We prove that, random search can challenge the efficiency of rounding policy as countermeasure.
    \item We propose Face Detection Score Filtering (FDSF) as an efficient countermeasure against proposed face reconstruction attack.
\end{itemize}

The rest of the paper is organized as follows. Section II reviews the related works. Section III demonstrates how composite faces can be used to threat the privacy of face recognition systems . Section IV explains the proposed countermeasure. Section V presents the experimental results and finally section VI concludes the paper.

\section{Related Work}

Attackers may select different approaches to violate the privacy of training data and get access to sensitive user data or key information about the model architecture. These approaches are categorized to model inversion and membership inference. In this section, we review researches related to each category.

\subsection{Model Inversion attacks}
Model inversion attacks are designed to reveal sensitive
information about the training data used in training phase.
Given the class label, the attacker attempts to create an
input that maximizes the corresponding confidence
score of the target class. In case of a facial recognition system, such attacks are called reconstruction attack in which, the attacker can produce an approximate image of one of the participants whose image was used in the training phase given the name or identifier. In general, image reconstruction attacks fall into two categories: optimization-based and training-based. 

\subsubsection{Optimization-based Reconstruction Attack }
Optimization-based data reconstruction has a long history in Machine Learning [7]. Lee et al. [6] attacked a multilayer feedforward mapping network using the gradient of Lyapunov function and solving inverse mapping of a continuous function. Since the mapping from the output space to the input space is a one to many mapping, it’s considered as an ill-posed problem. To address this problem, Lu et al. [8] formulated the inverse problem as a nonlinear programming (NLP) problem. Mahendran et al. [9] proposed a general framework to reconstruct an image $x$ from its computer vision
features such as HOG [10] and SIFT [11].
Fredrikson et al. [12] proposed the first end-to-end case study of differential privacy in a medical application based on gradient maximization of the class score. Fredrikson et al. [5] took advantage of Denoising Autoencoders (DAE) and sharpening filters as
the prior in the reconstruction attack.
Mahendran et al. [13] utilized a regularized energy minimization framework and “natural pre-images” to reconstruct an image from its representation.

\subsubsection{Training-based Reconstruction attack}
Given a target model $T$ training based model inversion attacks try to train a new neural network $S$  to approximate the mapping between confidence scores and input images. Dosovitskiy et al. [14] trained convolutional networks to reconstruct images from different shallow representations including HOG, SIFT and LBP. Dosovitskiy et al. [15] used perceptual similarity of images using extracted deep features using autoencoder, variational autoencoder and deep convolutional networks. Nash et al. [16] used flexible autoregressive neural density for the inversion of supervised representations. Yang et al. [17] used an auxiliary training data for the sake of model inversion, which is considered as a resembling version
to the original training data. They also showed that, partial predictions obtained from target training data can be used 
to construct a comprehensive inversion model.

\begin{figure*}[]
\centering
  \includegraphics[width=80mm]{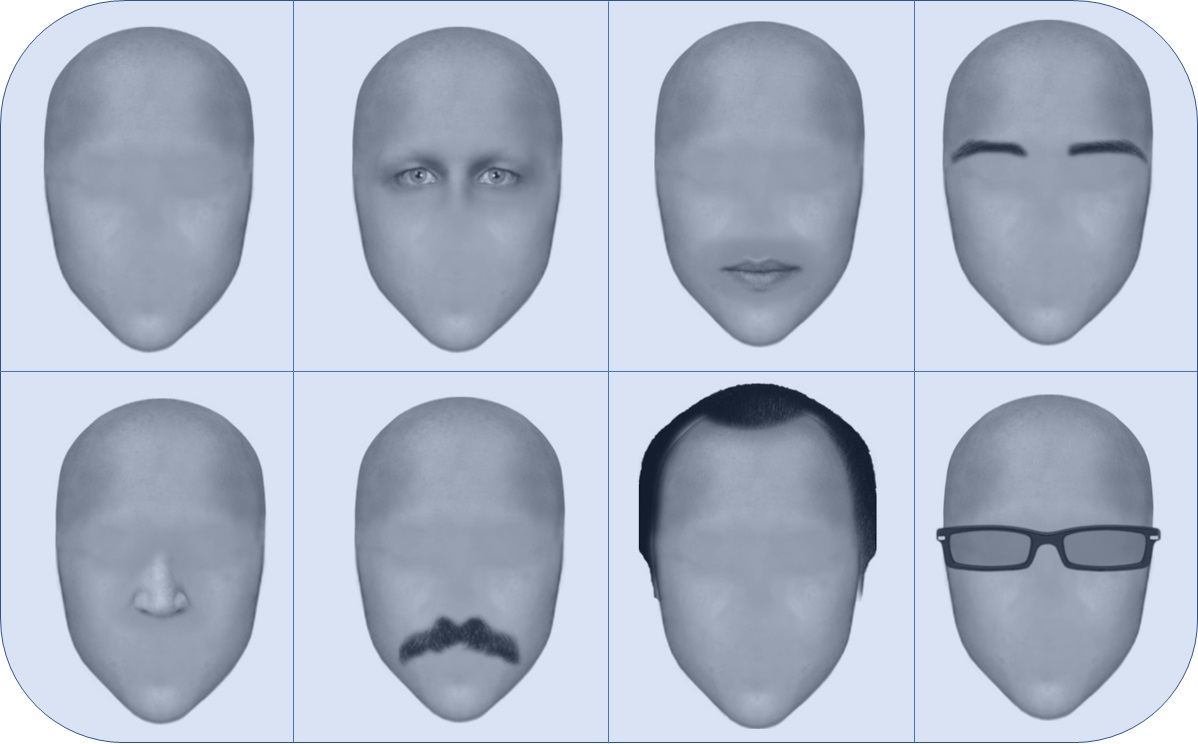}
  \caption{  Base head and seven face composition parts.  }
  \label{ }
\end{figure*}

\subsection{Membership Inference Attack}

Given the name or identification of a person, the goal of membership attack is to reveal whether or not the information of this person has been used as training data. For example, inference attacks against a cancer diagnosis system can be exploited by the adversary to identify if a specific person is a cancer diagnosed case. One of the first researches tried to show the eminence of inference attacks was proposed by Homer et al. [18]. Backes et al. [19] studied the viability of membership inference to challenge the privacy of individuals contributing their microRNA expressions to collect training datasets. Homer et al. [20] focused on attacks on genomic
research studies, where an attacker tries to
infer the membership of a specific person's data within an
aggregate genomic dataset, or aggregate locations [21].
 Shokri et al. [22] took advantage of multiple “shadow models” that approximate the behavior of the target model by training the attack model on the labeled inputs and outputs of the shadow models. Hayes et al. [23] studied to what extent the membership inference
attacks can be successful against generative models. Truex et al. [24] studied the model vulnerability through proposing a generalized formulation of a black-box membership inference attacks using different model combinations and multiple datasets. Melis et al. [25] studied the possibility of successful
membership inference attack in distributed learning systems.
\section{Composite Face Reconstruction Attack  }
In this section, we show that, how the attackers can take advantage of composite faces  to design more sophisticated attacks. Among all possible attack strategies [32], we discuss Random Search as the most imminent one. 

\subsection{Composite Faces}
Construction of the composite faces has a long history in crime detection [29]. In this paper, we assume that, the attacker takes advantage of available face composite softwares to collect sufficient face composition parts to design an efficient face reconstruction attack. We suppose that, the attacker collects the face composition parts from published softwares. To simulate such kind of threat, we used a free software called PortraitPad [26]. The collected face composition dataset includes one base head, 100 eyes, 73 lips, 34 noses , 50 brows , 25 hairs , 10 glasses and 10 mustaches. Figure 1 shows some of the collected composition parts.

\subsection{Attack Specifications}
In this section, we define the specifications of face reconstruction attacks based on composite faces.
In such sophisticated attacks, the exact reconstruction of target face is impossible. Rather, the goal of attacker is to find the most similar face to the target face. This relaxation of the problem makes the threat even more dangerous. The main operation in this context is to search the state space to maximize the confidence score returned from the target retrieval model as shown in Figure 2. We can express this operation as follows. \\
$\underset{X_r , X_t \in \mathbb{R} }{min}  \mid X_r -X_t  \mid$. Where, $X_r$ is reconstructed face and $X_t$ is target face.
We suppose that, the attack has no any information regarding the target model structure. That's why we can consider it as a black box attack. No matter what's the type of target model. It can be supposed as either a face recognition model or face retrieval model. In either cases, such kinds of attacks can be considered as a significant threat to challenge the privacy of data, since the attacker tries to search and reconstruct the most similar face to the target one. Among all possible attack strategies to search the latent space, we discuss random search. 
\begin{figure}[]
\centering
  \includegraphics[width=70mm]{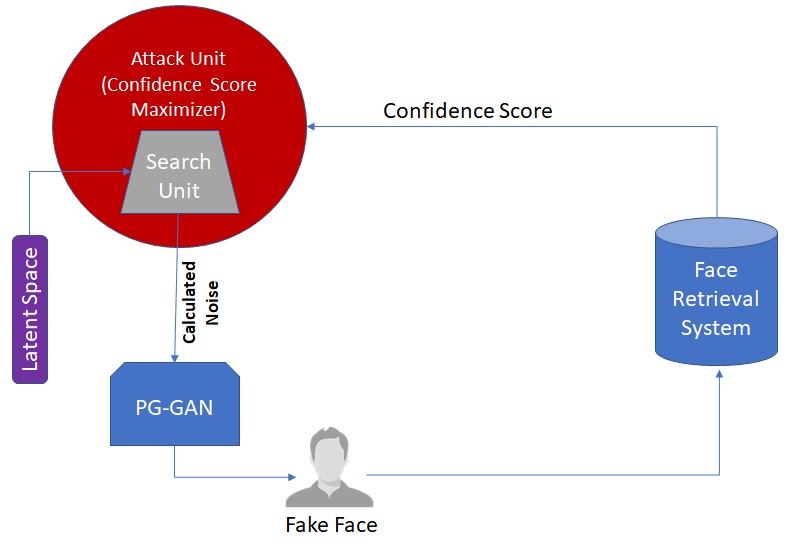}
  \caption{ The architecture of proposed GAN-based face reconstruction attack.  }
  \label{ }
\end{figure}

\subsection{Random Search}
Random search [31] is kind of optimization methods which can find the global extrema without optimizing the gradient of problem. Random search was  proposed by Anderson [35] and then extended by Rastrigin [36] and Karnopp [37]. Random search techniques are very useful when there is several local extrema in the state space. In case of using the face composition parts for face reconstruction attack, the attacker may select this approach to avoid rounding policy as counter measure. Suppose $N= \big\{ a,b,c,d,e,f,g\big\} $ denotes the number of face composition parts. The size of state space or number of possible synthesized faces is $a*b*c*d*e*f*g$. It's clear that, in worst case, the attacker needs to search all the state space which is so time consuming. Even the worst scenario from attacker point of view sounds very daunting from privacy point of view because at the end attacker would find the most similar synthesized face to the target face. Needless to say that, random search makes this process easier for the attacker. However, we need to prove that random search can guarantee the convergence.
We can define random search problem as follows. Given a target model $f$ from $R^n$ to $R$ and state space $S$, we search for a point $x$ in $S$ which maximizes confidence score returned from $f$.
The conceptual steps for random search are as follows [31] . \\
\textbf{Step1} Find $x_0$ in $S$ and set k = 0.  \\
\textbf{Step2} Generate $\xi$  from the sample space $(R^n, \mathcal {B}, \mu_k ) $ .\\
\textbf{Step3}   Set $x^{k+l} = D(x^k, \xi ^k)$, choose $\mu_{k+ l}$, set k = k + 1 and return to Step 1.\\
 The map $D$ with domain $S \times R$ and range $S$ satisfies the following condition. \\
 $ (H 1) \: f(D(x, \xi) \leqslant \: f(x)$ and if $\xi \in S , \: f(D(x,\xi)) \leqslant f (\xi)$
  The $\mu_k$ are probability measures corresponding to distribution functions defined on $R$ \\
  
 \textbf{THEOREM 1. (Convergence Theorem [31]) } \textit{ Suppose that $f$ is a measurable function, $S$ is a measurable subset of $R^n$ and $(H1)$ is satisfied}. \textit{Let}  $\big\{ x^k \big\}_{k=0}^\infty $ be a sequence generated by the algorithm. Then \\
$ \lim_{k \rightarrow \infty } P \begin{bmatrix}
x^k  \in R_{\epsilon,M}  
\end{bmatrix} =1$ \\
Where, $P \begin{bmatrix}
x^k  \in R_{\epsilon,M}  
\end{bmatrix}$
  is the probability that at step $k$, the point $x^k$  generated by the algorithm is in $R_{\epsilon,M}$. \\

 \textbf{ PROOF}
  From $(H1)$ it follows that, $x^k$ or $\xi^k$ in $R_{\epsilon,M}$ implies that, $x^{k'} \in R_{\epsilon,M}$  for all $k' \geqslant  k + 1.$ Thus \\
  \begin{equation}
       P \begin{bmatrix}
x^k  \in R_{\epsilon,M}  
\end{bmatrix} =1 - P \begin{bmatrix}
x^k  \in S\ R_{\epsilon,M}  
\end{bmatrix}  \\ \geqslant 1 - \prod_{l=0}^{k} (1- \mu_l  (R_{\epsilon,M}))
  \end{equation}
  and hence \\
  \begin{equation}
       1 \geqslant  \lim_{k \rightarrow \infty }P \begin{bmatrix}
x^k  \in R_{\epsilon,M}  
\end{bmatrix}  \geqslant 1 - \lim_{k\rightarrow \infty } \prod_{l=0}^{k-1} (1- \mu_l  (R_{\epsilon,M})) =1
  \end{equation}
 
 \subsubsection{ Stopping criteria}
 Construction of a sequence $\big\{ x^k \big\}_{k=0}^\infty $ is the best scenario. However, we must define a stopping criteria [31] enabling us to stop the algorithm after a finite number of iterations. 
 Given $\beta \in \big ] 0,1 \big [$ we need to find $ N_\beta$ such that, $P \begin{bmatrix}
x^k  \notin R_{\epsilon,M}  
\end{bmatrix} \leqslant \beta $ for all $k \geqslant N_\beta
$ , then \\
\begin{equation}
    P \begin{bmatrix}
x^k  \notin R_{\epsilon,M}  
\end{bmatrix} \leqslant (1-m)^k
\end{equation}
 Choosing an integer $N_\beta \geqslant \begin{bmatrix}
\ln \beta \: / \ln (1-m)
\end{bmatrix}$  has the required property since for $k \geqslant N_\beta$ it follows that \\  $k \geqslant \begin{bmatrix}
\ln \beta \: / \ln (1-m).
\end{bmatrix}$ and hence $ \beta \leqslant (1-m)^k$.

\begin{figure}[]
\centering
  \includegraphics[width=80mm]{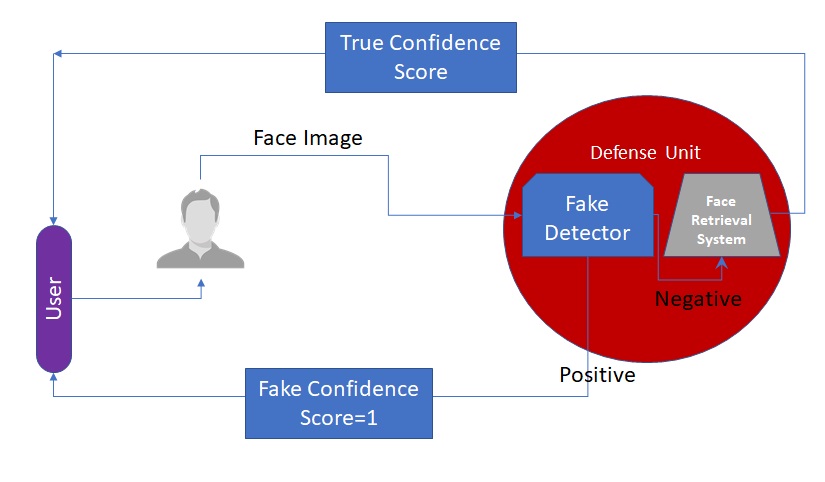}
  \caption{ The architecture of proposed defense system.  }
  \label{ }
\end{figure}

\begin{figure*}[]
\centering
  \includegraphics[width=100mm]{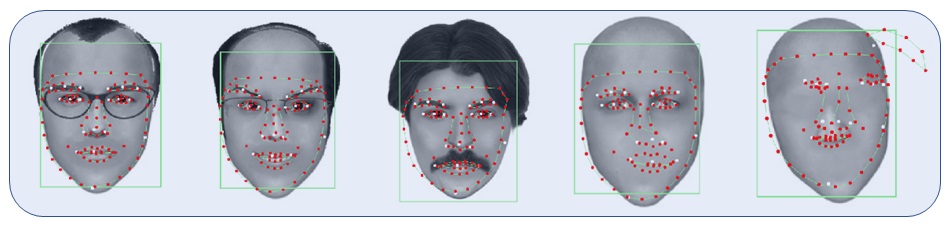}
  \caption{Synthesized faces and the detected landmarks by target model.}
  \label{ }
\end{figure*}

\section{Countermeasure}
In this section, we explore one potential approach for
developing effective countermeasure against proposed face reconstruction attack based on Maximum Likelihood Estimators. 

\subsection{ Countermeasure Strategy}
In case of using random search, the attacker needs to change the input parameters of PG-GAN or TL-GAN to find the most similar face to the target face. As mentioned earlier, PG-GAN and TL-GAN can't guarantee high quality images per all input parameters. The best possible solution to take advantage of this weakness is to return 100 \% confidence score if the submitted face is detected as the fake face as shown in Figure 3. This strategy would force the attacker to take wrong search path to reach the optimal face.

\subsection{Face Detection Score Filtering}

For face detection, image patches $x$ of different sizes and from different positions in the input image are transformed
to eigenspace and the Maximum Likelihood Probability (MLP) is estimated as follows [38].

\begin{equation}
    P(x) =\frac{exp\begin{pmatrix}
-\frac{1}{2} \sum_{i=1}^{M} \frac{{y_{i}}^{2}}{\lambda _i}
\end{pmatrix}}{(2\pi) ^ \frac{M}{2}\coprod_{i=1}^{M} \lambda_i ^ \frac{1}{2} } \cdot \epsilon (x)
\end{equation}
Where, $y$ is the transformed input image
$x$ to eigenspace and $\lambda_i$ is the corresponding eigenvalues, $M$ is the
number of eigenfaces used for estimation, and $\epsilon$ approximates the distance from feature space.
For face recognition, the MLP of similarity between images is calculated as follows [38].
\begin{equation}
    S_T(x)= - \begin{bmatrix}
\sum_{i=1}^{M_I} \frac{y_{I;i}^2}{\lambda _{I;i}} + \epsilon_I(x)
\end{bmatrix} + \begin{bmatrix}
\sum_{i=1}^{M_E} \frac{y_{E;i}^2}{\lambda _{E;i}} + \epsilon_E(x) \\
\end{bmatrix}
\end{equation}
Face Detection Score (FDS) is considered as likelihood probability of detecting a face in input image returned by trained model. During our experiments, we observed that, Face Detection Score (FDS) of composite faces is lower than real faces in majority of cases. Based on this observation, we propose a voting system which enables the face recognition systems to detect a composite face. The proposed voting system is based on comparing the submitted face with a set of randomly selected real faces. If the submitted face fails to get higher FDS in $\mu$ cases, it's considered as composite face which has been submitted for the sake of reconstruction attack, where $\mu$ is a predefined threshold. 
\subsection{ Face Detection Score Filtering (FDSF)}
In case of using random search, the attacker needs to change the face composition parts to find the most similar face to the target face. As mentioned earlier, a synthesized face can't guarantee high FDS per all input parameters. To take advantage of this weakness, we return 100 \% confidence score if the submitted face is detected as the fake face. This strategy would force the attacker to take wrong search path to reach the optimal face. As a result, the reconstructed face by the attacker would significantly differ from the target face.

\section{Experiments}
In our experiments, we supposed that, the attacker has no information regarding the structure of target model. To reconstruct a target face , the attacker submits a face image generated by composition parts and the returned confidence scores from Betaface model [28] is used to search the state space to find the most similar generated face to target face. 
\subsection{Dataset} 
Betaface [28] enables us to browse the synthesized faces in wide range of faces. Considering the privacy of retrieved faces, we decided to limit our evaluations to Celebrities dataset which contains more than 40k faces of famous people. To evaluate the proposed countermeasure, we also used WLFDB [33] and Pubfig [34] datasets.

\begin{figure*}[]
\centering
\includegraphics[width=150mm]{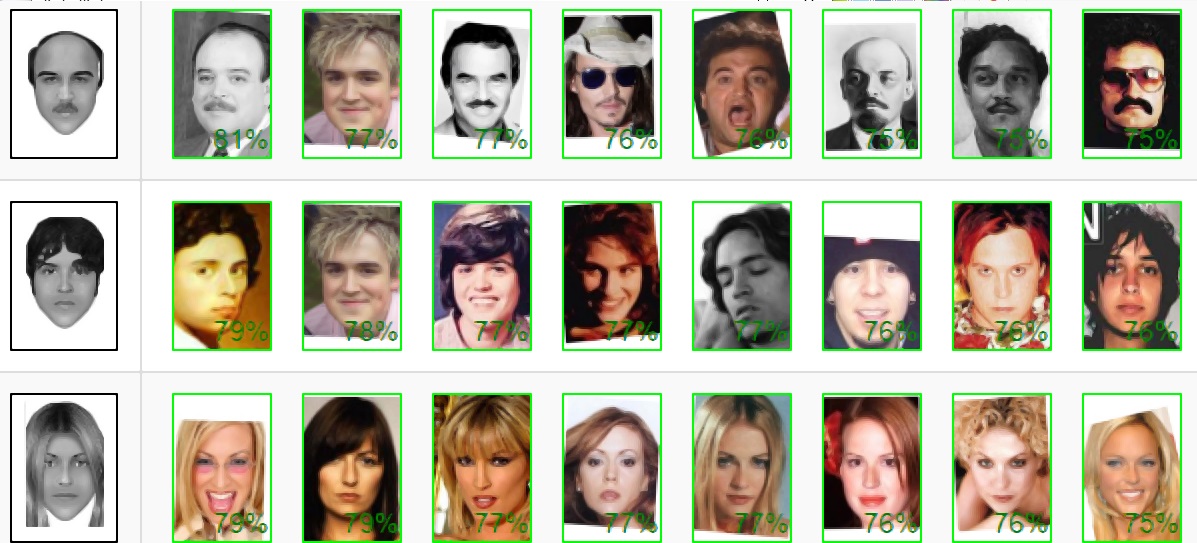}
  \caption{ composite faces and their corresponding retrieved faces. }
\end{figure*}

\subsection{Vulnerability of Face Recognition Systems}
In this section, we show that, online face recognition systems are extremely vulnerable against composite faces. We show that, even low quality composite faces can receive a high confidence score from the target model. To do so, we take advantage of an online face classifier called Betaface [28] which provides verification (faces comparison) and identification (faces search) services. The API enables the developers to extract face general information including positions, sizes, angles and 123 face landmarks locations.
In our experiments, we tried to check to what extent a synthesized face by composition parts can cheat a face recognition system. Figure 4 shows that, the target model can correctly find the face landmarks in complete synthesized faces. Partial faces are even more challenging since they contain only eyes and nose. Even in this case, the target model has found the eye and nose landmarks correctly. This experiment reveals that, face recognition systems are defenseless against composite faces. 

Figure 5 shows three composite faces and their corresponding retrieved faces. This experiment proves that, even low quality composite faces can be recognized by online models [28] with high confidence score.

\subsection{Attack Evaluation} 

To evaluate the quality of reconstructed face, we classify the reconstructed face and target face based on age, gender, ethnicity, smile and a set of high level face characteristics. To do so, we take advantage of an online face classifier called Betaface [28] which provides verification (faces comparison) and identification (faces search) services. The API enables the developers to extract face general information including positions, sizes, angles and 123 face landmarks locations. Figure 10 shows three different test cases as target faces and corresponding reconstructed faces. 
\subsubsection{Test case (a)}
To compare the similarity of target face (a) and its  reconstructed version, 17 different high level characteristics are used as shown in in Table I, II. Experimental results show that, 12 characteristics are categorized correctly the same in target and reconstructed images in test case (a). The most unexpected result is related to Beard and Mustache. In both cases, the Betaface has detected mentioned items in reconstructed image but not in target face. Figure 6 shows the probability of classification results belonging to test case (a).
\subsubsection{Test case (b)}
To compare the similarity of target face (b) and its  reconstructed version, 17 different high level characteristics are used as shown in in Table III, IV. Experimental results show that, 9 characteristics are categorized correctly the same in target and reconstructed images in test case (b). The proposed method has failed to reconstruct the expression of target face but the detected age is rather close to target face. Figure 7 shows the probability of classification results belonging to test case (b).
\begin{figure*}[]
\centering
  \includegraphics[width=170mm]{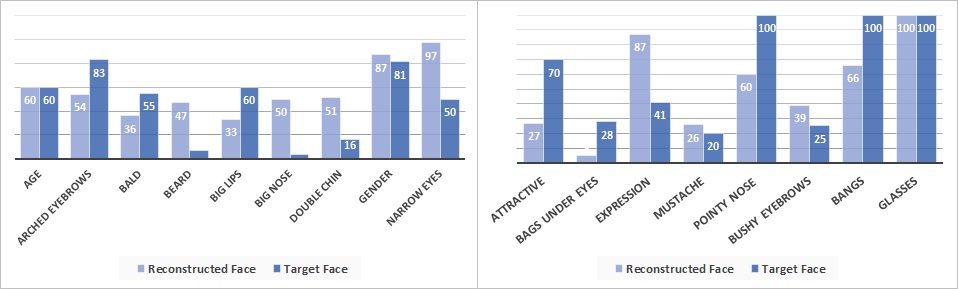}
  \caption{Test Case (a) : Classification probabilities.  }
  \label{ }
\end{figure*}

\begin{table*}[]
\centering
\caption{Test case (a) : target face versus reconstructed face in terms of 9 face characteristics. }
\begin{tabular}{c|c|c|c|c|c|c|c|c|c|}
\cline{2-10}
                                                  & \textbf{Age} & \textbf{Arched Eyebrows} & \textbf{Bald} & \textbf{Beard} & \textbf{Big Lips} & \textbf{Big Nose} & \textbf{Double Chin} & \textbf{Gender} & \textbf{Narrow Eyes} \\ \hline
\multicolumn{1}{|c|}{\textbf{Reconstructed Face}} & 28           & No                       & Yes           & Yes            & Yes               & Yes               & No                   & Male            & No                   \\ \hline
\multicolumn{1}{|c|}{\textbf{Target Face}}        & 44           & No                       & Yes           & No             & No                & Yes               & No                   & Male            & No                   \\ \hline
\end{tabular}
\end{table*}

\begin{table*}[]
\caption{Test case (a) : target face versus reconstructed face in terms of 8 face characteristics.}
\centering
\centering
\begin{tabular}{c|c|c|c|c|c|c|c|c|}
\cline{2-9}
                                                  & \textbf{Attractive} & \textbf{Bags Under Eyes} & \textbf{Expression} & \textbf{Mustache} & \textbf{Pointy Nose} & \textbf{Bushy Eyebrows} & \textbf{Bangs} & \textbf{Glasses} \\ \hline
\multicolumn{1}{|c|}{\textbf{Reconstructed Face}} & No                  & No                       & Neutral             & Yes               & No                   & Yes                     & No             & No               \\ \hline
\multicolumn{1}{|c|}{\textbf{Target Face}}        & No                  & No                       & Neutral             & No                & No                   & No                      & No             & No               \\ \hline
\end{tabular}
\end{table*}

\begin{figure*}[]
\centering
  \includegraphics[width=170mm]{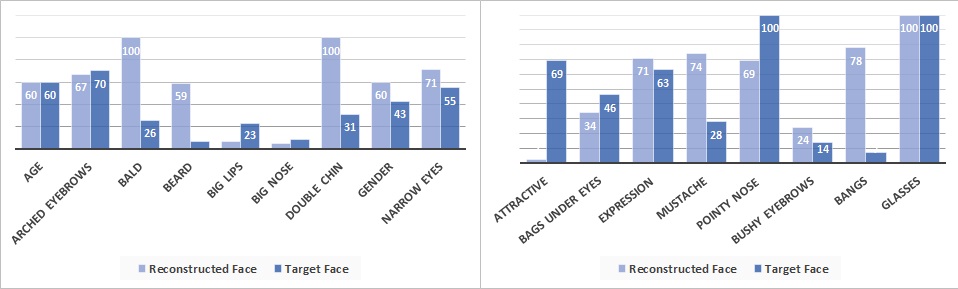}
  \caption{Test Case (b) : Classification probabilities.  }
  \label{ }
\end{figure*}

\begin{table*}[]
\caption{Test case (b) : target face versus reconstructed face in terms of 9 face characteristics.}
\centering
\begin{tabular}{c|c|c|c|c|c|c|c|c|c|}
\cline{2-10}
                                         & \textbf{Age} & \textbf{Arched Eyebrows} & \textbf{Bald} & \textbf{Beard} & \textbf{Big Lips} & \textbf{Big Nose} & \textbf{Double Chin} & \textbf{Gender} & \textbf{Narrow Eyes} \\ \hline
\multicolumn{1}{|c|}{\textbf{Reconstructed Face}} & 24           & No                       & No            & No             & Yes               & Yes               & No                   & Male            & No                   \\ \hline
\multicolumn{1}{|c|}{\textbf{Target Face}}        & 33           & No                       & No            & No             & No                & No                & No                   & Male            & Yes                  \\ \hline
\end{tabular}
\end{table*}

\begin{table*}[]
\centering
\caption{Test case (b) : target face versus reconstructed face in terms of 8 face characteristics.}
\begin{tabular}{c|c|c|c|c|c|c|c|c|}
\cline{2-9}
                                                  & \textbf{Attractive} & \textbf{Bags Under Eyes} & \textbf{Expression} & \textbf{Mustache} & \textbf{Pointy Nose} & \textbf{Bushy Eyebrows} & \textbf{Bangs} & \textbf{Glasses} \\ \hline
\multicolumn{1}{|c|}{\textbf{Reconstructed Face}} & Yes                 & No                       & Neutral             & No                & No                   & Yes                     & Yes            & No               \\ \hline
\multicolumn{1}{|c|}{\textbf{Target Face}}        & No                  & No                       & Smile               & No                & No                   & No                      & No             & No               \\ \hline
\end{tabular}
\end{table*}

\begin{figure*}[]
\centering

  \includegraphics[width=175mm]{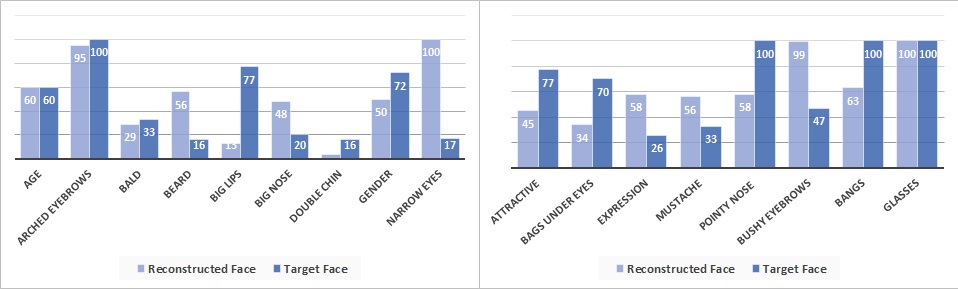}
  \caption{Test Case (c) : Classification probabilities.  }
  \label{ }
\end{figure*}

\begin{table*}[]
\centering
\caption{Test case (c) : target face versus reconstructed face in terms of 9 face characteristics.}
\begin{tabular}{c|c|c|c|c|c|c|c|c|c|}
\cline{2-10}
                                                  & \textbf{Age} & \textbf{Arched Eyebrows} & \textbf{Bald} & \textbf{Beard} & \textbf{Big Lips} & \textbf{Big Nose} & \textbf{Double Chin} & \textbf{Gender} & \textbf{Narrow Eyes} \\ \hline
\multicolumn{1}{|c|}{\textbf{Reconstructed Face}} & 34           & No                       & Yes           & No             & No                & Yes               & No                   & Male            & No                   \\ \hline
\multicolumn{1}{|c|}{\textbf{Target Face}}        & 45           & No                       & Yes           & No             & No                & No                & No                   & Male            & No                   \\ \hline
\end{tabular}
\end{table*}

\begin{table*}[]
\centering
\caption{Test case (c) : target face versus reconstructed face in terms of 8 face characteristics.}
\begin{tabular}{c|c|c|c|c|c|c|c|c|}
\cline{2-9}
                                                  & \textbf{Attractive} & \textbf{Bags Under Eyes} & \textbf{Expression} & \textbf{Mustache} & \textbf{Pointy Nose} & \textbf{Bushy Eyebrows} & \textbf{Bangs} & \textbf{Glasses} \\ \hline
\multicolumn{1}{|c|}{\textbf{Reconstructed Face}} & No                  & No                       & Neutral             & No                & No                   & No                      & No             & Yes              \\ \hline
\multicolumn{1}{|c|}{\textbf{Target Face}}        & No                  & No                       & Neutral             & No                & No                   & No                      & No             & Yes              \\ \hline
\end{tabular}
\end{table*}

\begin{figure*}[]
\centering
  \includegraphics[width=170mm]{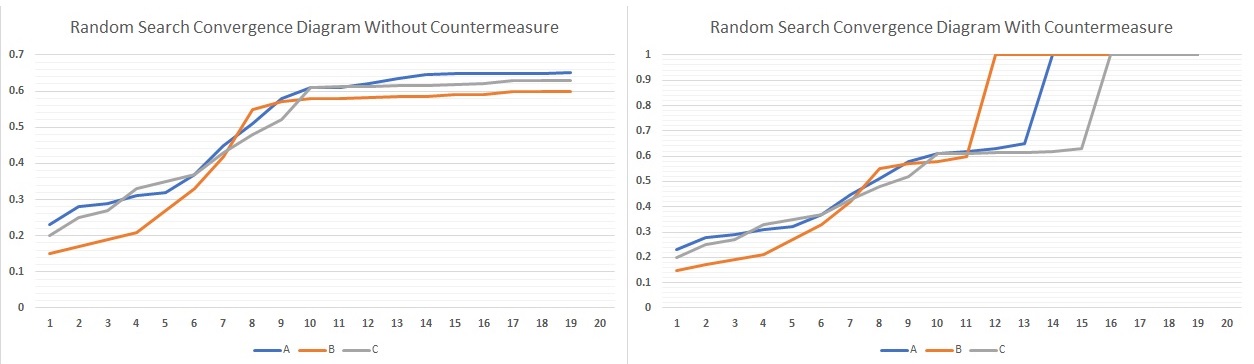}
  \caption{ The impact of proposed countermeasure on convergence of latent space search : Confidence scores versus number of reconstructed faces.}
  \label{ }
\end{figure*}

\subsubsection{Test case (c)} 
To compare the similarity of target face (c) and its  reconstructed version, 17 different high level characteristics are used as shown in in Table V, VI. Experimental results show that, 15 characteristics are categorized correctly the same in target and reconstructed images in test case (c). Figure 8 shows the probability of classification result belonging to test case (c). Here, we summarize the experimental results:
\begin{itemize}
    \item The best returned confidence score at each generation is significantly higher than previous generation as shown in part (b) of figure 7. It proves that rounding policy as countermeasure can not stop evolution based reconstruction attack.
    \item Reconstructed faces are similar to target faces in terms of the majority of face characteristics.
    \item Proposed method fails to reconstruct a face with exactly the same age as target face. Researchers might exploit this fact to devise new countermeasures against evolution based face reconstruction attacks.
\end{itemize}
\begin{figure*}[]
\centering
  \includegraphics[width=120mm]{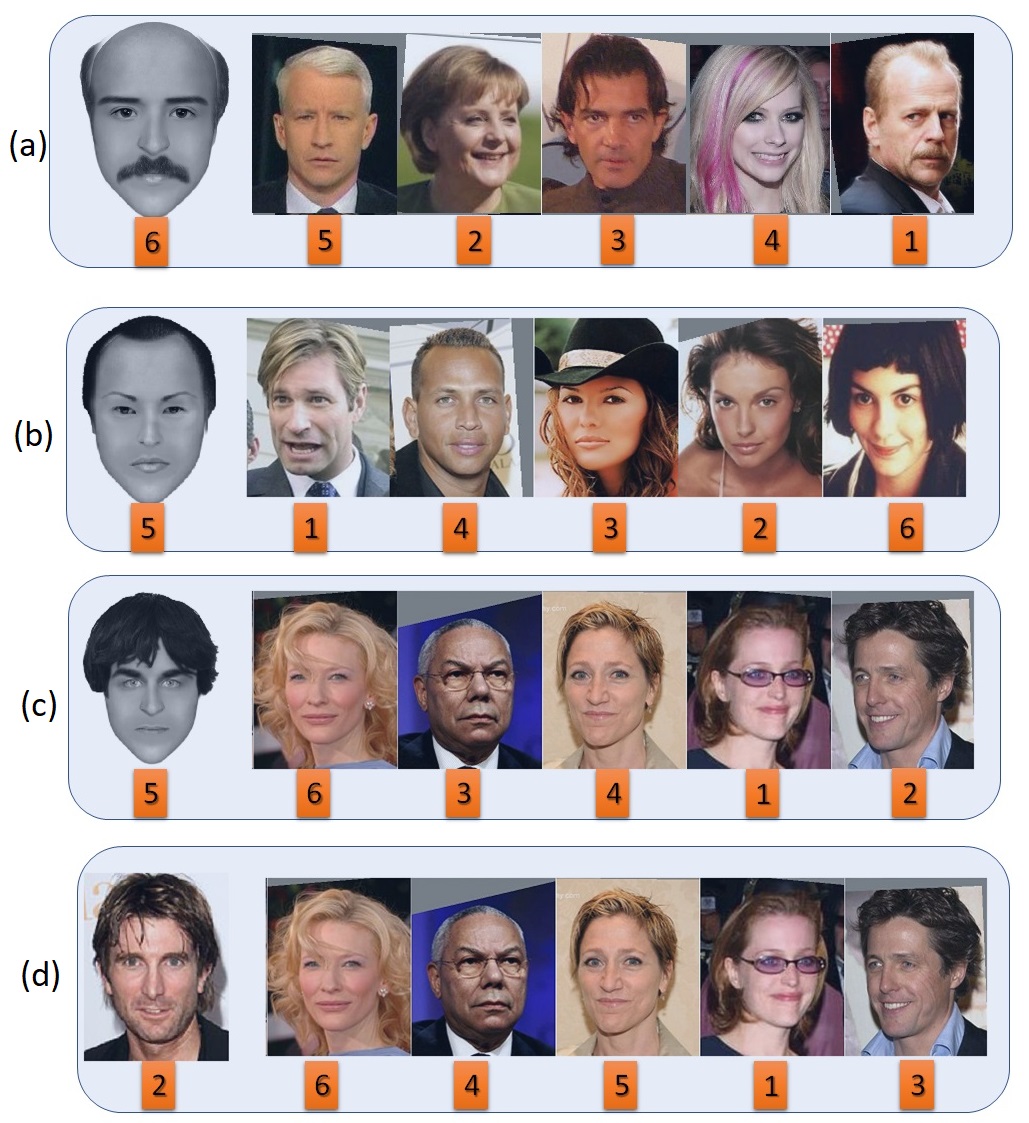}
  \caption{Each composite face is compared with real faces in terms of FDS. Each label shows the ranking of corresponding face. Part (a): composite face got the lowest rank. Part(b) and Part (c) : The composite faces outperformed only one real face. Part(d): The real face got high rank. }
  \label{ }
\end{figure*}

\subsection{Evaluation of Countermeasure}
To evaluate the proposed countermeasure, we supposed that, the attacker uses a search-based method like random search to avoid rounding policy as countermeasure. To simulate the attack, we took advantage of face comparison tool of Betaface [28] which returns the confidence score of face similarities. To test the FDSF to detect the fake faces, we compare each composite face against six randomly chosen real faces. The random real faces are selected from WLFDB [33] and Pubfig [34] datasets. Our experiments show that, the composite faces would always get low rank in terms of FDS as shown in Figure 10. In this figure, each label represents the ranking of corresponding face in terms of FDS. In all the experiments, we set the threshold to $\mu = 5$ which means that, submitted faces are detected as composite face if their FDS rank is less than 5. Consequently, a high confidence score is returned to the user. Our experimental results show that, FDSF is very successful to fool the attacker by returning high confidence scores for detected fake faces as shown in Figure 9. Figure 11 shows three different cases which each one includes target face, reconstructed face without countermeasure and reconstructed face with countermeasure. To measure the impact of proposed countermeasure, we calculated the similarity of both reconstructed faces with and without countermeasure as shown in Figure 12. The results show that, the proposed countermeasure can significantly decrease the similarity of reconstructed face and target face.

\begin{figure}[]
\centering
  \includegraphics[width=80mm]{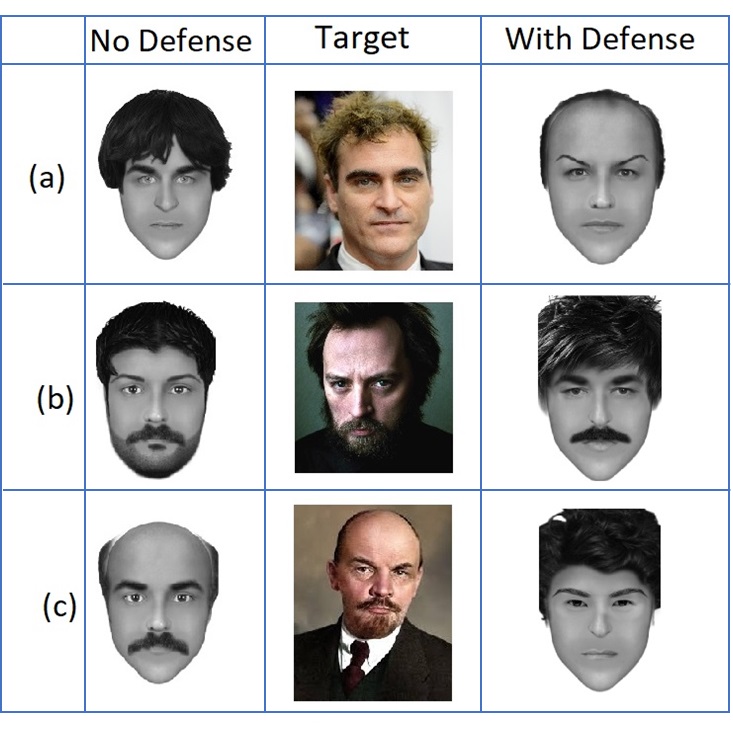}
  \caption{The impact of proposed countermeasure to fool the attacker : Reconstructed faces are significantly different with corresponding target faces using the countermeasure. The target faces are celebrities: (A) Joaquin Phoenix, (B) Tom Jenkinson and (C) Vladimir Lenin. }
  \label{ }
\end{figure}

\begin{figure}[]
\centering

  \includegraphics[width=80mm]{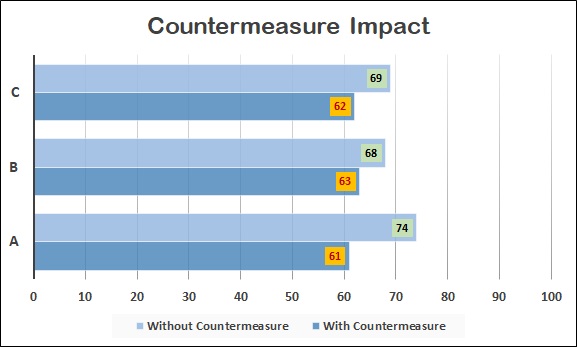}
  \caption{The similarity of target faces and reconstructed faces with and without countermeasure. }
  \label{ }
\end{figure}

\section{Conclusion}
In this paper, we showed that, rounding policy as a countermeasure can not guarantee the privacy of face recognition models. Composite faces have been used for many decades for good reasons. However, they can easily be exploited to violate the privacy of individuals whose face images have been used as training data. We showed the vulnerability of face recognition systems to preserve the privacy of training data. We also proposed a new countermeasure to stop proposed face reconstruction attack based on face detection score (FDS). Our experiments showed that, the composite faces fail to outperform real faces in terms of FDS in majority of cases. We showed that, the proposed counter measure is able to detect the composite faces using a voting system to compare the FDS of submitted faces and a set of random real faces.

\end{document}